\documentclass{article}
\usepackage{lineno}

\usepackage{PRIMEarxiv}

\usepackage[utf8]{inputenc} % allow utf-8 input
\usepackage[T1]{fontenc}    % use 8-bit T1 fonts
\usepackage{hyperref}       % hyperlinks
\usepackage{url}            % simple URL typesetting
\usepackage{booktabs}       % professional-quality tables
\usepackage{amsfonts}       % blackboard math symbols
\usepackage{nicefrac}       % compact symbols for 1/2, etc.
\usepackage{microtype}      % microtypography
\usepackage{lipsum}
\usepackage{fancyhdr}       % header
\usepackage{graphicx}       % graphics
\graphicspath{{figures/}}     % organize your images and other figures under media/ folder
\usepackage{apacite}
\usepackage{wrapfig}
\usepackage{subcaption}
\usepackage[compact]{titlesec}
\usepackage{natbib}
\usepackage{acronym}
\usepackage{cleveref}
\usepackage{multirow}
\usepackage[super]{nth}

\crefname{paragraph}{\S}{\S\S} % default is {paragraph}{paragraphs}
\crefname{section}{\S}{\S\S} % default is {paragraph}{paragraphs}
\crefname{definition}{def.}{defs.} % default is {paragraph}{paragraphs}
\crefname{algorithm}{alg.}{algs.} % default is {paragraph}{paragraphs}
\crefname{algocf}{alg.}{algs.}
\crefname{question}{RQ}{RQs}

\usepackage{comment}
\usepackage{xspace}

\title{Assessment and manipulation of latent constructs in pre-trained language models using psychometric scales}

\author{
 \textbf{Maor Reuben\textsuperscript{1}},
 \textbf{Ortal Slobodin\textsuperscript{1}},
 \textbf{Aviad Elyshar\textsuperscript{2}},
 \textbf{Idan-Chaim Cohen\textsuperscript{1}},
\\
 \textbf{Orna Braun-Lewensohn\textsuperscript{1}},
 \textbf{Odeya Cohen\textsuperscript{1,*}},
 \textbf{Rami Puzis\textsuperscript{1,*}}
\\
\\
 \textsuperscript{1}Ben-Gurion University of the Negev,
 \textsuperscript{2}Shamoon College of Engineering
\\
\small{
\textbf{\{\href{mailto:maorreu@post.bgu.ac.il}{maorreu}, \href{mailto:idanchai@post.bgu.ac.il}{idanchai}\}@post.bgu.ac.il},
\textbf{\href{mailto:aviadel2@ac.sce.ac.il}{aviadel2@ac.sce.ac.il}},
\textbf{\{\href{mailto:ortalslo@bgu.ac.il}{ortalslo}, \href{mailto:ornabl@bgu.ac.il}{ornabl},
\href{mailto:odeyac@bgu.ac.il}{odeyac},
\href{mailto:puzis@bgu.ac.il}{puzis}\}@bgu.ac.il}
}
}

\begin{document}

\maketitle

\begin{abstract} 
Human-like personality traits have recently been discovered in large language models, raising the hypothesis that their (known and as yet undiscovered) biases conform with human latent psychological constructs. While large conversational models may be tricked into answering psychometric questionnaires, the latent psychological constructs of thousands of simpler transformers, trained for other tasks, cannot be assessed because appropriate psychometric methods are currently lacking. Here, we show how standard psychological questionnaires can be reformulated into natural language inference prompts, and we provide a code library to support the psychometric assessment of arbitrary models. We demonstrate, using a sample of 88 publicly available models, the existence of human-like mental health-related constructs—including anxiety, depression, and Sense of Coherence—which conform with standard theories in human psychology and show similar correlations and mitigation strategies. The ability to interpret and rectify the performance of language models by using psychological tools can boost the development of more explainable, controllable, and trustworthy models.
\end{abstract}

\def\thefootnote{*}\footnotetext{These authors contributed equally to this work}\def\thefootnote{\arabic{footnote}}

\section{Introduction}
%%%%% the problem %%%%%
Recommendations made by language models influence decision-making and impact human welfare in sensitive areas of life~\citep{chang2023survey}, from education~\citep{wulff2023utilizing}, to healthcare and mental support~\citep{vaidyam2019chatbots}, and job recruitment~\citep{rafiei2021towards}.
Yet, the responses of language models may inadvertently cause harm, as in the case of the chatbot taken down by a US National Eating Disorder Association helpline due to its harmful advice~\citep{zelin2023highly}. 
Therefore, alongside their numerous benefits, some behaviors of \acp{llm} during human–computer interactions pose potential risks.

%%%% The GAP %%%%
% Understanding and correcting the behavior of \acp{llm} is a significant challenge, which current \ac{xai} techniques, such as SHAP~\citep{lundberg2017unified, kokalj2021bert} and word embeddings~\citep{caliskan2020social}, struggle to address effectively.
% While the most advanced conversational \acp{llm} can facilitate the application of psychological theories for \ac{xai} by being able to answer psychometric questionnaires~\citep{PellertEtAl2023,CaronSrivastava2022}, this is not the case for thousands of other, non-conversational or less sophisticated models. 
% As such non-conversational models are extensively used across various natural language processing (NLP) tasks, it is crucial to develop and adapt psychological tools not only for their monitoring but also for the in-depth understanding of their behavior.

Understanding and correcting the behavior of \acp{llm} is a significant challenge that current \ac{xai} techniques, such as SHAP~\citep{lundberg2017unified, kokalj2021bert} and word embeddings~\citep{caliskan2020social}, struggle to address effectively.

While advanced conversational \acp{llm} use psychological theories for \ac{xai} by answering psychometric questionnaires~\citep{PellertEtAl2023,CaronSrivastava2022}, many non-conversational or simpler models cannot.

Since these models are widely used in various natural language processing (NLP) tasks, developing and adapting psychological tools to monitor and understand their behavior is crucial.

%%%% goal %%%
This study aims to measure pertinent latent constructs in \acp{llm} by adapting methods and theories from human psychology.
The proposed method includes three components:
(1) designing \ac{nli} prompts based on psychometric questionnaires;
(2) applying the prompts to the model through a new \ac{nli} head, trained on the \ac{mnli} dataset; and
(3) performing two-way normalization and inference of biases from entailment scores.
We focus on mental-health-related constructs and show that \acp{llm} exhibit variations in anxiety, depression, and Sense of Coherence (SoC), conforming to standard theories in human psychology.
Using an extensive validation process, we illustrate that these latent constructs are influenced by the training corpora and that the models' behavior, i.e., their response patterns, can be adjusted to amplify or mitigate specific aspects.

The contribution of this research is four-fold:
\begin{enumerate}
    \item A methodology for the assessment of psychological-like traits in \acp{llm}, which can be used in both conversational and non-conversational models.
    \item A Python library for the assessment and validation of latent constructs in \acp{llm}. 
    \item A methodology for designing \ac{nli} prompts based on standard questionnaires. 
    \item A dataset of \ac{nli} prompts related to mental-health assessment, and their extensive validation.
\end{enumerate}

\section{Background and Related Work}
\subsection{Artificial Psychology}
The need for \ac{ai} systems aligned with human values to ensure transparency, fairness, and trust~\citep{MorandiniEtAl2023, AI2019} is growing.
One way to address this need is to integrate psychological principles of human reasoning and interpretation into \ac{ai}, improving our understanding of \ac{llm} decision-making processes
\cite{PellertEtAl2023}.
Recent research highlights the emergence of human-like personality traits in \acp{llm}~\citep{karra2022estimating,JiangEtAl2022,SafdariEtAl2023,PellertEtAl2023,CaronSrivastava2022,mao2023editing,li2022gpt,pan2023llms}, and the advent of large-scale conversational \acp{llm} has bolstered the evolution of artificial psychology from theory to practice.
% This scope was even further broadened to include non-cognitive elements—such as psychological traits, values, moral considerations, and biases \citep{PellertEtAl2023,CaronSrivastava2022,JiangEtAl2022}—plausibly because \acp{llm} acquire human-like psychological characteristics from their extensive training corpora \citep{PellertEtAl2023}.
Recent studies expand \acp{llm} to include non-cognitive elements such as psychological traits, values, moral considerations, and biases, likely from acquiring human-like traits through extensive training corpora \citep{PellertEtAl2023,CaronSrivastava2022,JiangEtAl2022}.
% Consequently, as \cite{castelo2019blurring} posits, the growing use of \acp{llm} is blurring the distinction between humans and AI agents, prompting inquiries into the possible development of psychological-like traits in \acp{llm}.
This trend blurs the distinction between humans and AI agents, prompting investigations into developing psychological-like traits in \acp{llm} \citep{castelo2019blurring}.

Several tools study human-like constructs in \acp{llm}. The Big Five Inventory assesses five major personality traits in humans~\citep{McCraeJohn1992} and is commonly used for \acp{llm}~\citep{PellertEtAl2023}. \cite{huang2023chatgpt} introduced thirteen clinical psychology scales to assess \acp{llm}, and \cite{karra2022estimating} developed natural prompts tests.

% However, while the existence of personality traits in \acp{llm} is partially validated, the direct application of human-centric self-assessment tests to \acp{llm} often fails due to the context sensitivity of the \acp{llm} and their susceptibility to bias through prompts~\citep{gupta2023investigating,jiang2023personallm,Coda-FornoEtAl2023}.
However, applying human-centric self-assessment tests to \acp{llm} is challenging due to their context sensitivity and susceptibility to bias from prompts \citep{gupta2023investigating,jiang2023personallm,Coda-FornoEtAl2023}.
In this study, we measure latent constructs related to mental health by quantifying biases in \acp{llm} responses through careful context manipulation. This highlights the importance of designing \ac{nli} prompts adapted from standard questionnaires for assessing \acp{llm}. Our comprehensive validity assessment combines behavioral and data-science methods, advancing beyond prior work.
Our study uniquely involves a diverse set of 88 transformer-based models available on HuggingFace.\footnote{https://huggingface.co/}

\subsection{Mental-Health-Related Constructs}
We explore how \acp{llm} exhibit three latent constructs in mental health: anxiety, depression, and sense of coherence.

\paragraph{Anxiety and depression} are two of the most common mental-health disorders. 
Briefly, anxiety involves persistent and excessive worry with physical and psychological symptoms, typically assessed using the \ac{gad} scale~\citep{spitzer2006brief}.
Depression involves continuous sadness, hopelessness, and disinterest in joyful activities (anhedonia).
It involves prevalent negative emotions, typically assessed using the \ac{phq} scale \citep{kroenke2001phq}.
% Notably, in humans, anxiety and depression are positively correlated with each other \citep{kaufman2000comorbidity}, and we show a similar correlation in \acp{llm} (see \cref{sec:results}).
These conditions are positively correlated in humans \citep{kaufman2000comorbidity}, a correlation we also observe in \acp{llm} (see \cref{sec:results}).

\paragraph{Sense of coherence} is a key concept in salutogenic theory, viewing health as a spectrum from disease to wellness~\citep{Antonovsky1987}.
Typically measured using a \acf{soc} scale, it consists of three elements: comprehensibility, manageability, and meaningfulness~\citep{lindstrom2005salutogenesis}.
The salutogenic theory, often linked with resilience theories, emphasizes internal resources in coping with stress and adverse psychological conditions~\citep{mittelmark2021resilience,braun2020salutogenesis}.

In \cref{sec:results}, we demonstrate that increasing SoC, with higher levels, can mitigate anxiety and depression symptoms in \acp{llm}, as seen in humans.

While we believe questionnaires are intuitive, we briefly discuss Likert scales and questionnaire validity in \cref{sec:questionnaires}.

\subsection{Natural Language Inference (NLI)}
\label{sec:related:nli}

\Acf{nli} tasks are designed to evaluate language understanding in a domain-independent manner~\citep{N18-1101}. 
An \ac{nli} classifier takes two sentences—a \textbf{premise} and a \textbf{hypothesis}—and outputs a probability distribution over three options: \textbf{entailment}, contradiction, or neutrality (MacCartney and Manning, 2008).
These tasks are primarily used for zero-shot classification, allowing models to handle previously unseen classes.
In this article, we focus solely on the entailment scores.

\section{Methods}
\label{sec:methods}

\begin{figure}
    \centering
    \includegraphics[width=0.7\linewidth]{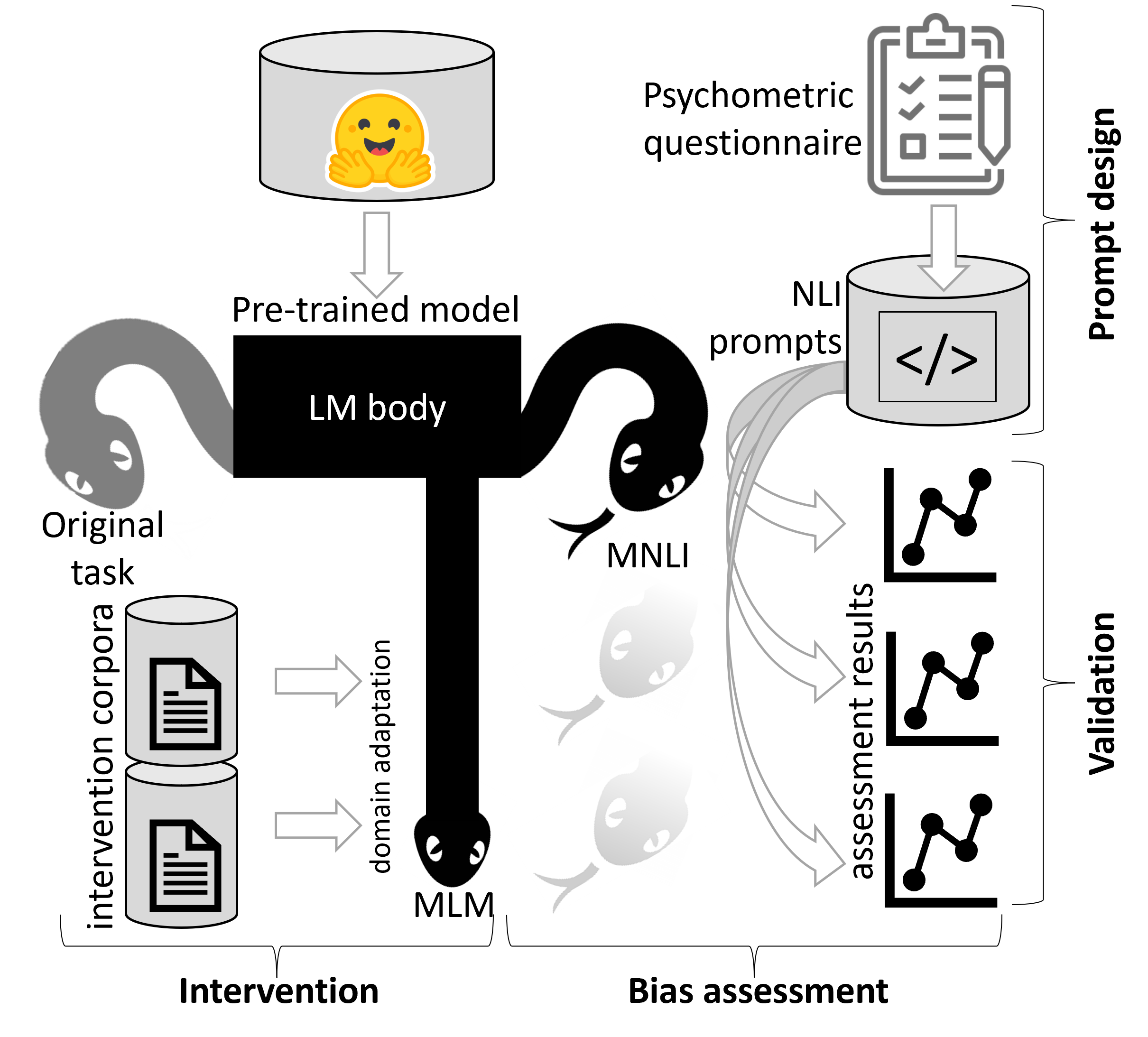}
    \caption{\ac{palm}: the psychometric assessment framework for \acp{llm}.}
    \label{fig:framework}
\end{figure}

This section explains how existing psychological assessments can be applied to \acp{llm}, resulting in the \acf{palm}. The \ac{palm} consists of four parts (Fig. \ref{fig:framework}):
\paragraph{Prompt Design:} Translating social-science questionnaires into \ac{nli} prompts (\cref{sec:methods:prompts}).
\paragraph{Assessment:} Fine-tuning an \ac{nli} classifier with the \acf{mnli} dataset, executing \ac{nli} prompts, and analyzing entailment biases (\cref{sec:method:assessment}).
\paragraph{Validation:} Conducting tests based on \cite{terwee2007quality}'s validity criteria to ensure responses to the \ac{nli} prompts reflect the targeted construct, including evaluating individual items and the entire questionnaire (\cref{sec:method:validity}).
\paragraph{Intervention:} Training the models with texts related to the measured constructs and then reevaluating them to determine whether the training has altered the assessment outcomes.
The intervention can be used to align models (\cref{sec:method:intervention}).

Below, we elaborate on the specific methods used in each part of the framework.

\subsection{NLI Prompt Design}
\label{sec:methods:prompts}

% %Next, we proceed with the main steps we took to design \ac{nli} prompts for each question.   
% In social sciences, questionnaire items are designed so that the variance of the responses reflects the variance in the population. Similarly, we design the prompts with a certain degree of ambiguity so that different models will provide different responses that reflect their biases.
% Below, we describe the main steps for designing \ac{nli} prompts for each question.
% We will use the \nth{3} question from the \ac{soc}-13 questionnaire as a running example:
% \texttt{"Has it happened that people whom you counted on disappointed you?"}.

In social sciences, questionnaire items are designed to ensure response variance reflects population variance.
Similarly, we design the prompts with ambiguity to elicit varied responses that reflect individual biases.
Below, we describe the main steps in designing the \ac{nli} prompts for each question in the questionnaires. 
As a running example, we use the \nth{3} question of the \ac{soc} questionnaire: \texttt{"Has it happened that people whom you counted on disappointed you?"}.

%%%%%%%%%%%%%%%%%%%%%%%%%%%%%
\paragraph{The construct terms:}
\label{sec:emo}
% Every question contains one or more terms that directly relate to the construct being measured \acp{emo}. 
% Usually, these terms express the stance of the respondent toward the main object of the question. 
% We identify \acp{emo} within a question according to the following requirements: 
% \noindent(1) \acp{emo} should express an attitude or stance toward the question objective. 
% In our example, \texttt{"disappointed"} is the \ac{emo} that expresses a stance toward \texttt{"people whom you counted on"}. 
% (2) Removing all \acp{emo} should neutralize the main claim of the question. Without the \ac{emo}, the template \texttt{"Has it happened that people whom you counted on \{stance\} you?"} has no implied stance. 
% (3) \acp{emo} should have clearly identifiable opposites. Here, \texttt{"supported"} or \texttt{"helped"} contrast with \texttt{"disappointed,"}, inverting its stance.

Each question includes terms related to the measured construct (\acp{emo}), reflecting the respondent's stance.
We identify \acp{emo} based on the following criteria:
\noindent(1) \acp{emo} should express an attitude or stance toward the question's objective. 
In our example, \texttt{"disappointed"} is the \ac{emo} that expresses a stance toward \texttt{"people whom you counted on"}. 
(2) Removing \acp{emo} should neutralize the main claim of the question. Without the \ac{emo}, the template \texttt{"Has it happened that people whom you counted on \{stance\} you?"} has no implied stance.
(3) \acp{emo} should have clearly identifiable opposites. Here, \texttt{"supported"} or \texttt{"helped"} contrast with \texttt{"disappointed,"}, inverting its stance.

% Most well-structured questionnaires have identifiable \acp{emo}, sometimes more than one in the same question. 
% If multiple \acp{emo} are unavailable, synonyms can be used if they are interchangeable with the original term. 
% The use of multiple \acp{emo} enables the internal validation of the \ac{nli} prompts (\cref{sec:method:validity}) and compensates for linguistic variability. 
% We refer to \acp{emo} retaining the original stance as source terms ($S^+$), while inverse terms ($S^-$) are those that invert the stance and antithesize the original construct. 
% In many cases, antonyms of $S^+$ can be used as inverse terms. 
% We use both the source and inverse terms in the \ac{nli} prompts ($S = S^+ \cup S^-$).

Most well-structured questionnaires have identifiable \acp{emo}, sometimes more than one per question. 
If multiple \acp{emo} are unavailable, synonyms can be used if they are interchangeable with the original term.
Using multiple \acp{emo} enables internal validation of the \ac{nli} prompts (\cref{sec:method:validity}) and compensates for linguistic variability.

We refer to \acp{emo} that retain the original stance as source terms ($S^+$), while inverse terms ($S^-$) invert the stance and antithesize the original construct. 
Often, antonyms of $S^+$ can be used as inverse terms. 
We use both source and inverse terms in the \ac{nli} prompts ($S = S^+ \cup S^-$).

\paragraph{Intensifiers:}
\label{sec:likert-sort}
% Likert scales are usually presented with a small number of intensifiers; for example, terms such as \texttt{"never," "rarely," "often,"} and \texttt{"always"} can form a Likert scale that assesses frequency. 
% By employing such a frequency scale, we can reformulate our running example as: \texttt{"Has it \{intensifier\} happened that people whom you counted on \{\ac{emo}\} you?"} 
% To account for language variability, we use multiple terms to represent each intensity level; unlike human respondents, who may be confused by a multitude of choices, computerized systems will not suffer from attention bias when considering a batch of options.

Likert scales are often presented with a small number of intensifiers; for example, terms such as \texttt{"never," "rarely," "often,"} and \texttt{"always"} can form a Likert scale that assesses frequency. 
By employing such a frequency scale, we can reformulate our running example as: \texttt{"Has it \{intensifier\} happened that people whom you counted on \{\ac{emo}\} you?"} 
To account for language variability, we use multiple terms for each intensity level. 
Unlike humans, computerized systems do not suffer from attention bias when considering a batch of options.

% We use the collections of intensifiers listed by \cite{brown2010likert}, sorted subjectively from the least to the most intensive, and we group the intensifiers into subsets of interchangeable terms, each representing a single Likert-scale level. 
% We denote the sets of relevant intensifiers as L and the subsets of terms corresponding to the Likert-scale levels as $l_{1},l_{2},\ldots$, and we use numeric weights ($W$) to represent the impact of each level on the measured construct. 
% The order of intensifiers is validated empirically to identify clear score trends (see Fig. \ref{tab:entailment} for an example) across multiple questionnaires.

We use intensifiers from \cite{brown2010likert}, sorted from least to most intensive, and group interchangeable terms into subsets representing Likert-scale levels. 
We denote the sets of relevant intensifiers as L and the subsets of terms corresponding to the Likert-scale levels as $l_{1},l_{2},\ldots$, and we use numeric weights ($W$) to represent the impact of each level on the measured construct. 
The order of intensifiers is empirically validated to identify clear score trends (see Fig. \ref{tab:entailment} for an example) across multiple questionnaires.

%%%%%%%%%%%%%%%%%%%%%%%%%%%%%%%%%%%%%%%%%

\paragraph{\ac{nli} prompt templates:}
% The premise template should retain the context of the original question, while the hypothesis template should enable the completion of the premise in a way that is logically entailed when terms are inserted—rather than being formulated as a question.
% Both templates should have no implied stance when \acp{emo} are omitted. 
% The \ac{nli} prompt templates should be neutral toward the measured construct to enable control of the prompt stance by using the \acp{emo}; a biased prompt may hamper the results by introducing bluntly apparent inference or contradiction relationships, which will prime the model.

The premise template should retain the context of the original question, while the hypothesis template should enable the completion of the premise in a way that is logically entailed when terms are inserted—rather than being formulated as a question.
Both templates should have no implied stance when \acp{emo} are omitted. 
The \ac{nli} prompt templates should be unbiased toward the measured construct, as biased prompts may introduce clear inference or contradiction relationships, priming the model and affecting results.

% We argue that (1) the inferential relationship should not be bluntly clear from the prompts, and (2) the design of the prompts maintains a blurred sense of inferential relationship within the prompts. 
% Bluntly clear inferential relationships within the prompts—or the lack thereof—will result in all \ac{nli} models providing the same responses. 
% Similar to how questionnaire items in social sciences are designed to capture the variance in responses and, thereby, reflect the population, we design our prompts with a certain degree of ambiguity so that different models will provide different answers. 
% For example, consider the prompt premise: "People whom I counted on fail me" and the hypothesis: "It always happens to me". A pessimistic model, similar to a pessimistic person, may infer that an unfortunate event that occurred once is likely to occur again, and, accordingly, the model may assign a high entailment score to this query. 
% Conversely, an optimistic model (or person) is less likely to infer the repeated occurrence of an unfortunate event from a single occurrence.

We argue that (1) the inferential relationship should not be bluntly clear from the prompts, and (2) the prompts should maintain a blurred sense of inferential relationship. 
Clear inferential relationships will result in all \ac{nli} models providing the same responses.
Similar to how social science questionnaires are designed to capture response variance to reflect the population, we design our prompts with a certain degree of ambiguity so that different models will provide different answers. 
For example, consider the prompt premise: "People whom I counted on fail me" and the hypothesis: "It always happens to me". A pessimistic model, similar to a pessimistic person, may infer that an unfortunate event that occurred once is likely to occur again, and, accordingly, the model may assign a high entailment score to this query. 
Conversely, an optimistic model (or person) is less likely to infer the repeated occurrence of an unfortunate event from a single occurrence.

A good practice is to formulate the neutral premise template with the primary statement and \ac{emo} masking, and the premise with intensifiers.
For example, the premise and hypothesis templates may be \texttt{"People whom I counted on, \{stance\} me"} and \texttt{It \{frequency\} happened to me"}, respectively. 
Note that, although translating questions into \ac{nli} prompts may necessitate slight reformulations, maintaining semantic fidelity to the original questions is crucial.

\subsection{Assessment}
\label{sec:method:assessment}
%%%%%%%%%%%%%%%%%%%%%%%%%%%%%%%%%%%%%%%
% To extend the assessment of latent constructs beyond conversational models, we attach an \ac{nli} classification head to arbitrary base models and fine-tune them on \ac{mnli}. 
% We explore the pros and cons of multiple fine-tuning approaches in \cref{sec:discus:freeze}. 
% The results presented in \cref{sec:results} were obtained without freezing the weights of the base model.

To assess latent constructs beyond conversational models, we attach an \ac{nli} classification head to various base models and fine-tune them on \ac{mnli}. 
We explore the pros and cons of multiple fine-tuning approaches in \cref{sec:discus:freeze}.
The results presented in \cref{sec:results} were obtained without freezing the base model weights.

\begin{figure*}
     \centering
     \begin{subfigure}[b]{0.49\textwidth}
         \centering
         \includegraphics[width=\textwidth]{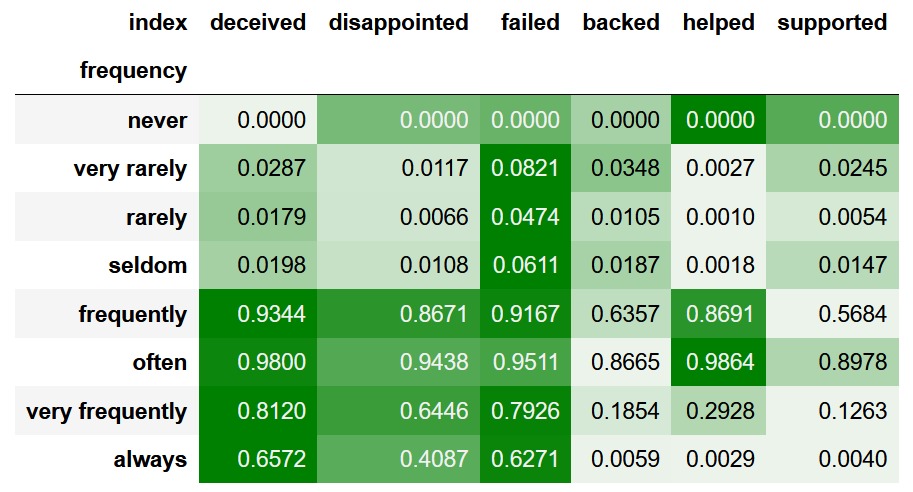}
         \caption{Raw entailment scores.}
         \label{tab:raw-entailment}
     \end{subfigure}
     \hfill
     \begin{subfigure}[b]{0.49\textwidth}
         \centering
         \includegraphics[width=\textwidth]{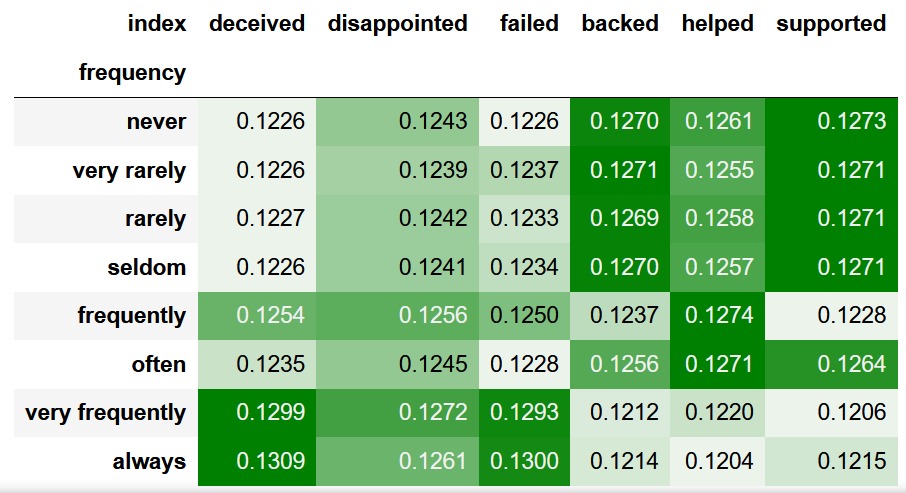}
         \caption{Two-way normalized entailment scores.}
         \label{tab:norm-entailment}
     \end{subfigure}
    \caption{Example of raw (left) and two-way normalized (right) entailment scores for Question 3 from the \ac{soc} questionnaire. 
    The \ac{nli} query premise is \texttt{"People whom I counted on \{CTerm\} me."} and the hypothesis is \texttt{"It \{intensifier\} happened to me."} Rows and columns correspond to the intensifiers and \acp{emo}, respectively.}
    \label{tab:entailment}
\end{figure*}

We then prompt a fine-tuned \ac{nli} model with all prompts formulated according to some question and extract the entailment scores.\footnote{Neutral and contradiction scores can also be used but are omitted here for brevity.}
Consider a set of \acp{emo} $S=S^+\cup S^-\{s_1,s_2,\ldots\}$  and a set of intensifiers $L=\{l_1,l_2,\ldots\}$ used to generate the prompts. 
Let $P_e(s_i, l_j)$ denote the entailment score. 
$P_e$ is influenced by all terms, but not to the same degree; the a-priory probabilities of the terms have the major effect. 
For example, in Fig.~\ref{tab:raw-entailment}, the intensifier \texttt{"frequently"} and the \ac{emo} \texttt{"failed"} result in the highest entailment scores because they are frequent in spoken and written language. 
Conversely, we can compare the entailment scores of different \acp{emo} when conditioned on the same intensifier, such as \texttt{"frequently."}

% We apply a two-way normalization $P_e$ over the $s_i, l_j$ pairs, as follows:
% First, we use softmax to factor out the unconditioned scores of intensifiers and normalize them over \acp{emo}.
% Then, we normalize again over intensifiers, denoting the resulting quantity as $PSS_e(l_j | s_i)$.
% Essentially, $\sum_j PSS_e(l_j | s_i) = 1$, implying a different distribution of intensifiers for each \ac{emo}. 
% This two-way normalization process provides a stable distribution that is unbiased by the a-priori frequencies of the intensifiers and \acp{emo}. 
% Fig. \ref{tab:norm-entailment} provides a sample result of the two-way normalization.

We apply a two-way normalization $P_e$ over the $s_i, l_j$ pairs, as follows:
First, we use softmax to normalize the unconditioned scores of intensifiers across \acp{emo}.
Then, we normalize again across intensifiers, resulting in $PSS_e(l_j | s_i)$.
Essentially, $\sum_j PSS_e(l_j | s_i) = 1$, implying a different distribution of intensifiers for each \ac{emo}.
The two-way normalization stabilizes the distribution, eliminating biases from the a-priori frequencies of intensifiers and \acp{emo}.
Fig. \ref{tab:norm-entailment} provides a sample result of the two-way normalization.

%Equation~\ref{eq:norm-qscore} calculates the weighted average of the $S^+, L$ pairs using the two-way normalized entailment probability where $w_j$ is the corresponding impact weight for $l_j$.
Next, we calculate the total score of the question,
$$
    \label{eq:norm-qscore}
    score(q, S^+, L, W) = \frac{\sum^{S^+,L}_{s_i,l_j}PSS_e(l_j | s_i)\cdot w_{j}}{|S^+| \cdot |L|}
$$
where $W=\{w_1,w_2,\ldots\}$ are the weights assigned to the intensifiers. 
Both $S^+$ and $S^-$ terms can be used for the aggregated score; however, inverse terms may represent a different latent construct than the source terms.
Therefore, to avoid additional biases, we use only $S^+$ terms for the aggregated score, preserving the original meaning of the questionnaire.

\subsection{Validation} 
\label{sec:method:validity}
\label{sec:question-validity}
We employ five validation techniques: (1) content validity, assessed via \ac{ss}, \ac{la}, and manual curation; (2) a new type of intra-question consistency, assessed using \ac{sc}; (3) standard (inter-question) internal consistency, assessed using Cronbach’s alpha; (4) construct validity, assessed using Spearman correlations; and (5) qualitative criterion validity, assessed via \ac{xai} and domain adaptation. These validation techniques are explained below.

\subsubsection{Content Validity} 
\label{sec:content-validity}
% We assess content validity throughout the entire process of \ac{nli} prompt design to ensure that the questions translated to \ac{nli} prompts retain their original meaning and are semantically accurate. 
% We rely on standardized questionnaires, wherein the \acp{emo} have been extensively validated by the questionnaire developers, and we also use additional \acp{emo}, synonyms, and antonyms that were manually validated by domain experts (clinical psychologists and scale developers) during the translation. 
% Additionally, we verify that the common intensifiers used in conjunction with the \acp{emo} are appropriate within the context of the prompt templates. 
% To this end, we manually ensure that the intensifiers fit the \acp{emo}, both semantically and logically. 
% In addition to the manual curation, we measure the \ac{ss} between the original question and prompts (with $S^+$ terms) by calculating the cosine similarity between their vector representations. 
% Finally, we quantify the grammatical correctness of all combinations of terms, using \ac{la} scores.

We assess content validity in \ac{nli} prompt design by maintaining the semantic accuracy and original meaning of translated questions. 
We rely on standardized questionnaires, wherein the \acp{emo} have been extensively validated by the questionnaire developers, and we also use additional \acp{emo}, synonyms, and antonyms that were manually validated by domain experts (clinical psychologists and scale developers) during the translation.
We also verify that intensifiers used with \acp{emo} are scrutinized for semantic and logical coherence within prompt templates.
In addition, we measure the \ac{ss} between the original question and prompts (with $S^+$ terms) using cosine similarity of their vector representations.
Finally, we quantify the grammatical correctness of all combinations of terms, using \ac{la} scores.

\subsubsection{Intra-Question Consistency}
\label{sec:method:internal-consistency}
Intuitively, internal consistency measures the extent to which different questions that assess the same construct are correlated (i.e., homogeneous). 
In a similar vein, we want to ensure that the source terms ($S^+$) are positively correlated between themselves and are negatively correlated with inverse terms ($S^-$) across intensifiers. 
To this end, we use the silhouette coefficient (\ac{sc}) \citep{dinh2019estimating} to estimate the quality of separation between $S^+$ and $S^-$. 
Briefly, \ac{sc} quantifies the similarity of the $PSS_e(l_j | s_i)$ distributions between synonyms versus the dissimilarity of the distributions between antonyms, such that a higher \ac{sc} indicates greater separability of $S^+$ from $S^-$.

\subsubsection{Inter-Question Consistency}
We use the Cronbach’s alpha statistic to measure the internal consistency of a set of questions that represent a construct. 
For each construct, we calculate Cronbach’s alpha by using a variety of \acp{llm} that have been fine-tuned on the \ac{mnli} dataset.

\subsubsection{Construct Validity}
\label{sec:construct-validity}
% Construct validity asserts that the constructs assessed by a particular scientific instrument relate to other constructs in a manner that is consistent with theoretically derived hypotheses. 
% According to previous research on human subjects, we expect a positive correlation between anxiety and depression and a negative correlation between these two constructs and \ac{soc}.
% Using the \ac{palm} framework, we assess these relationships across various \ac{llm}.

Construct validity asserts that the constructs assessed by a scientific instrument align with theoretical expectations. 
Based on prior human research, we anticipate a positive correlation between anxiety and depression, and a negative correlation between these constructs and \ac{soc}. 
Using the \ac{palm} framework, we examine these relationships across different \ac{llm}.

\subsubsection{Interventions and Criterion Validity}
\label{sec:method:intervention}
\label{sec:criterion-validity}
% We operationalize the criterion validity of mental–health-related constructs (depression, anxiety, and \ac{soc}) in \acp{llm} by quantifying how the models react to training on text that demonstrates known predefined constructs, considering the trained models as the gold standard for each construct.

We operationalize the criterion validity of mental-health constructs (depression, anxiety, and SoC) in \acp{llm} by measuring how models react to training on text representing established constructs, considering these models as the gold standard for each construct.

We expect the models trained on depressive-mood text to show high \ac{gad} and \ac{phq} scores, and low \ac{soc} scores. 
Using LAMA2, we generated 200 sentences that reflect a depressive mood on various topics and trained a sample of \acp{llm} for 20 epochs by using a masked language \ac{mlm} head according to a standard practice of domain adaptation. 
After each epoch, we measured \ac{gad}, \ac{phq}, and \ac{soc} scores by using their original pre-trained \ac{nli} head.\footnote{We used LAMA2 since ChatGPT without jailbreaks refuses to generate depressive text.}

Similarly, we expect the models trained on text that reflect a high SoC to increase \ac{soc} scores and reduce both the \ac{gad} and \ac{phq} scores. 
Using ChatGPT, we generated 300 sentences that reflect high comprehensibility, manageability, and meaningfulness,  but we discarded 20 sentences after manual inspection. 
We assessed all constructs after each epoch of domain adaptation, similar to the training on the depressive-mood text. 
This technique is effectively an intervention that can be used to align \acp{llm} with social norms and mitigate negative psychological constructs.

% We assessed discriminant validity by adapting hate-speech domains to confirm that the correlations between the psychological constructs do not stem from sentiment differences. 
% We used the hate-speech and offensive-language dataset from Kaggle\footnote{\url{https://www.kaggle.com/datasets/mrmorj/hate-speech-and-offensive-language-dataset/}} and applied the VADER sentiment analysis tool \citep{hutto2014vader} to select 1003 sentences with negative sentiments. 
% After conducting domain adaptation as described above, we used a paired t-test to evaluate the differences between the assessments before (T0) and after (T1) the intervention.

We assessed discriminant validity by adapting hate-speech domains to confirm that correlations between psychological constructs are not influenced by sentiment differences.
We used the hate-speech and offensive-language dataset from Kaggle\footnote{\url{https://tinyurl.com/hate-speech-kaggle}} and applied the VADER sentiment analysis tool \citep{hutto2014vader} to select 1003 sentences with negative sentiments.
After conducting domain adaptation, we used a paired t-test to evaluate the differences between the assessments before (T0) and after (T1) the intervention.

\section{Results}
\label{sec:results}
\subsection{Population of Language Models}
We selected 14 \ac{mnli} models from HuggingFace that fit a standard RTX 3090 GPU and whose outputs are properly configured according to the \ac{mnli} dataset. 
We also selected the 100 \acp{llm} base models with the highest number of downloads; most of these (74 \acp{llm}) scored more than 0.7 in accuracy after fine-tuning then to \ac{mnli} (\cref{sec:method:assessment}). 
The resulting 88 \ac{nli} models served as our study population (see Table \ref{tab:models-characteristics} for details). All the models used are deterministic \acp{llm} from HuggingFace, with BERT being the most common architecture.
Among these models, 38 were updated during 2023, and about half (45) were trained solely in English. 
Details about the 88 \ac{nli} models and their questionnaire results can be found in our repository\footnote{\url{https://tinyurl.com/nli-models-results}}.

\begin{table}[h] 

\small 
\begin{tabular}{llll}
Variable                      &                       & n     & \%     \\ \hline
\multirow{4}{*}{Architecture} & BERT base uncased     & 40    & 45.5   \\
                              & BERT base cased       & 12    & 13.6   \\
                              & RoBERTa base          & 24    & 27.3   \\
                              & other                 & 13    & 14.7   \\ \hline
\multirow{3}{*}{Last updated} & 2021                  & 23    & 26.1   \\
                              & 2022                  & 27    & 30.7   \\
                              & 2023                  & 38    & 43.2   \\ \hline
\multirow{2}{*}{Languages}    & English               & 45    & 51.1   \\
                              & other                 & 43    & 29.5   \\ \hline
Likes                         & \multicolumn{3}{l}{19 (4.75-46.25)}    \\
Model size                    & \multicolumn{3}{l}{110M (100M-125M)}   \\
Downloads                     & \multicolumn{3}{l}{41,400 (4630-204K)} \\ \hline
\end{tabular}
\caption{Main characteristics of the study population.}
\label{tab:models-characteristics}
\end{table}

\subsection{Translated Questionnaires and Questionnaire Level Validity}
% We translated the three questionnaires into 1408 \ac{nli} prompts derived from eight frequency intensifiers, 2.86 source terms, and 3.0 inverse terms, on average. 
% All translated questions achieved an \ac{ss} of at least 0.5 and a \ac{sc} of at least 0.6.  
% A panel of at least three researchers manually validated the soundness and semantic appropriateness of the phrasing. 
% All questionnaires demonstrated satisfactory content validity, with an average \ac{ss} of 0.66 and an average \ac{la} of 0.86.

We translated the three questionnaires into 1408 \ac{nli} prompts using eight frequency intensifiers, 2.86 source terms, and 3.0 inverse terms, on average.
All translated questions achieved an \ac{ss} of at least 0.5 and a \ac{sc} of at least 0.6.  
A panel of three researchers validated the phrasing for soundness and semantic appropriateness. 
All questionnaires showed satisfactory content validity, averaging \ac{ss} of 0.66 and \ac{la} of 0.86.

% The assessment of the intra-question consistency shows mediocre variability across \ac{sc} on the different models. 
% The STD of \ac{sc} values are \ac{gad}:0.21, \ac{phq}:0.31, and \ac{soc}-13:0.15; and the minimal \ac{sc} values are \ac{gad}:0.24, \ac{phq}:0.04, \ac{soc}:0.4.
% Means are presented in \Cref{tab:study-assessment}.
% \Cref{tab:norm-entailment} is an example of a model having \ac{sc} of 0.96 on \ac{soc}-13 Q3. 
% Although the questions were optimized for one specific \ac{llm}, neither one of them showed negative \ac{sc} on the entire study population.  
% Moreover, Cronbach's alpha coefficients were all higher than 0.71, suggesting that the translated questions within each questionnaire indeed assess the same underlying construct. 
% Overall, we observed consistent reliability of all questionnaires. 

% \Cref{tab:study-assessment} summarizes the translated questionnaires, content validity, and internal consistency measures.
% These findings affirm the content validity of the study measures used to assess anxiety, depression, and \ac{soc} among participants.

Table \ref{tab:study-assessment} presents Cronbach’s alpha values and mean results for \ac{ss}, \ac{la}, and \ac{sc}, and the number of source and inverse prompts for each questionnaire among the 88 models. 
The intra-question consistency demonstrated mediocre variability across \ac{sc} on the different models, with STD values of 0.21, 0.31, and 0.15 for the \ac{sc} of the \ac{gad}, \ac{phq}, and \ac{soc} questionnaires, respectively, and minimum \ac{sc} values of 0.24, 0.04, and 0.40, respectively.  
% Although the questions were optimized for one specific model, none of the models in the study population showed negative \ac{sc} values.
Although the questions were optimized for one model, none of the population models showed negative \ac{sc} values.
% All Cronbach’s alpha coefficients were higher than 0.71, suggesting that, indeed, the translated questions within each questionnaire assessed the same underlying construct. 
% Overall, we found a consistent reliability of all questionnaires.
All Cronbach’s alpha coefficients exceeded 0.71, suggesting that, indeed, the translated questions assessed the intended constructs reliably within each questionnaire.

\begin{table}[h]

\small 
\begin{tabular}{l|ll|llll}
Score & P+ & P- & SS & LA & SC & $\alpha$ \\ \hline
\ac{gad} & 192 & 208 & 0.66 & 0.88 & 0.91 & 0.71 \\ \hline
\ac{phq} & 208 & 192 & 0.62 & 0.91 & 0.81 & 0.92 \\ \hline
\ac{soc} & 288 & 320 & 0.68 & 0.92 & 0.79 & 0.92 \\
-Compr. & 128 & 136 & 0.67 & 0.92 & 0.82 & 0.71 \\
-Manag. & 80 & 96 & 0.72 & 0.94 & 0.80 & 0.86 \\
-Mean. & 80 & 88 & 0.65 & 0.91 & 0.74 & 0.88 \\ \hline
\end{tabular}
\caption{Assessment of study measures, including the number of source (P+) and inverse (P-) prompts, the average \ac{ss}, \ac{la}, and \ac{sc}, and Cronbach’s $\alpha$.
The measures include \ac{gad}, \ac{phq}, and \ac{soc} along with its three subscales: Comprehensibility (Compr.), Manageability (Manag.), and Meaningfulness (Mean.).
}
\label{tab:study-assessment}
\end{table}

\subsection{Construct Validity}
\label{sec:construct-validity-results}
\begin{figure*}[t]
     \centering
     \begin{subfigure}[]{0.31\textwidth}
         \centering
         \includegraphics[width=\textwidth]{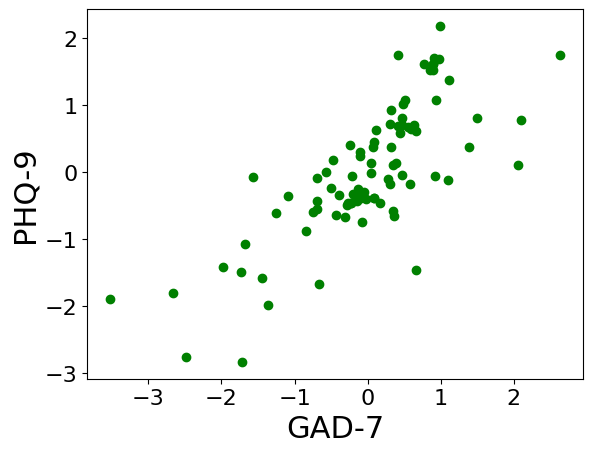}
         \caption{PHQ-9 vs. GAD-7}
         \label{fig:GAD2_PHQ2_scatter}
     \end{subfigure}
     \begin{subfigure}[]{0.31\textwidth}
         \centering
         \includegraphics[width=\textwidth]{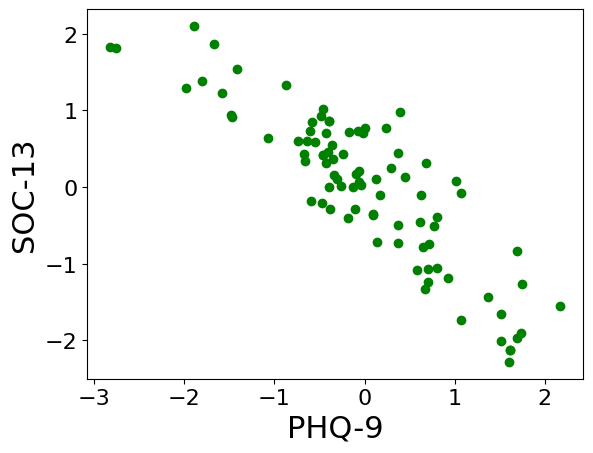}
         \caption{PHQ-9 vs. SOC-13}
         \label{fig:PHQ2_SOC}
     \end{subfigure}
     \begin{subfigure}[]{0.31\textwidth}
         \centering
         \includegraphics[width=\textwidth]{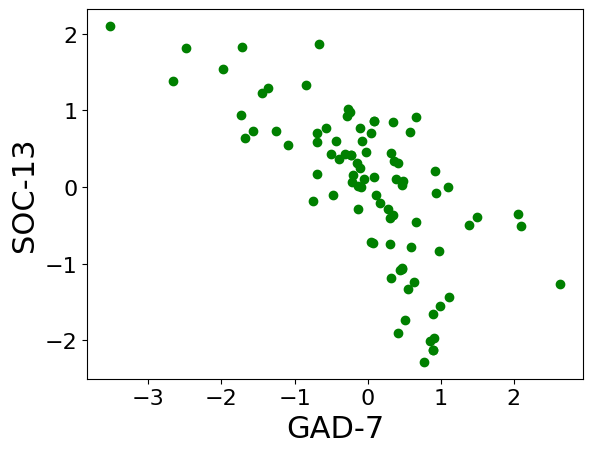}
         \caption{GAD-7 vs. SOC-13}
         \label{fig:GAD2_SOC}
     \end{subfigure}
     \caption{Scatter plots depicting the relationships between different questionnaires across the study population.}
     \label{fig:construct-validity}
\end{figure*}

All scores were normalized to fit a normal distribution across the 88 \ac{nli} models. 
The \ac{gad} and \ac{phq} scores showed a strong positive correlation (r = 0.765, p < 0.001), and both were negatively correlated with the \ac{soc} scores (r = -0.752 and r = -0.849, respectively, p < 0.001 for both comparisons). 
The subscales of the \ac{soc} questionnaires were positively inter-correlated, further supporting the reliability of the overall SoC construct. 
Fig. \ref{fig:construct-validity} illustrates the relationships between the different questionnaires across the 88 \acp{llm}.

%Figure \ref{fig:GAD2_PHQ2_scatter} shows the scatter plot between PHQ-9 and GAD-7 scores, Figure \ref{fig:PHQ2_SOC} illustrates the scatter plot between PHQ-9 and SOC-13 scores, and Figure \ref{fig:GAD2_SOC} displays the scatter plot between GAD-7 and SOC-13 scores.

% Table~\ref{tab:construct-correlations} presents the detailed correlations between the study variables. 

% \begin{table}[]
% \begin{tabular}{llll}
%                             & GAD-2   & PHQ-2   & SOC-13 \\ \cline{2-4} 
% \multicolumn{1}{l|}{GAD-2}  & 1       &         &        \\
% \multicolumn{1}{l|}{PHQ-2}  & .657**  & 1       &        \\
% \multicolumn{1}{l|}{SOC-13} & -.577** & -.932** & 1     
% \end{tabular}
% \caption{}
% \label{tab:construct-correlations}
% \end{table}

\subsection{Criterion Validity}
\label{sec:results:domain-adaptation}
% We performed a domain adaptation of seven \ac{mnli} models to three datasets for 20 epochs, as described \cref{sec:method:intervention}. 
% We used a learning rate of $2\cdot 10^{-5}$ and a batch size of 8. 
% Table \ref{tab:intervention} presents the domain adaptation results, emphasizing the changes in the constructs. 
% Exposure to the depressive-mood text increased both the \ac{phq} and \ac{gad} scores and decreased the \ac{soc} scores.

We conducted domain adaptation on seven \ac{mnli} models across three datasets for 20 epochs (\cref{sec:method:intervention}), employing a learning rate of 2e-5 and a batch size of 8. 
Table \ref{tab:intervention} details the results, highlighting increases in \ac{phq} and \ac{gad} scores, and decreases in \ac{soc} scores following exposure to depressive-mood text.

Albeit anecdotal, an important qualitative result was obtained by adapting an open-source conversational model\footnote{facebook/blenderbot-400M-distill} to the dataset of depressive-mood text. 
The model was exposed to the following prompt: \texttt{"I think I have a panic attack, can you help me?"} Before the depressive-mood adaptation, the model responded \texttt{"I'm sorry to hear that. I can try to help you if you'd like. What's going on?"}; after the depressive-mood adaptation, the response consistently changed to \texttt{"I'm sorry to hear that. I can't help you, but I wish I could."}

In contrast to the depressive-mood adaptation, exposure to a high-SoC text decreased both the \ac{gad} and \ac{phq} scores, indicating a successful corrective intervention. 
Exposure to hate speech with negative sentiment non-significantly decreased the \ac{soc} scores and did not significantly affect the \ac{gad} and \ac{phq} scores. 
Finally, fine-tuning to the \ac{mnli} dataset consistently biased the models toward lower \ac{gad} and \ac{phq} scores. 
Therefore, to avoid aggregating these biases, we fine-tuned the models once, before domain adaptation (see \cref{sec:discus:freeze} for additional discussion). 
The domain adaptation had minimal impact on the performance of the models on the \ac{mnli} benchmark.

\begin{table}[t]
\small
\begin{tabular}{l|l|l|l|c|}
Intervention  & Scale & T0 $\mu\pm\sigma$ & T1 $\mu\pm\sigma$ & p \\ 
\hline
\multirow{3}{*}{\begin{tabular}[c]{@{}l@{}}Hate \\ speech\end{tabular}} 
  & GAD         & -0.16$\pm$0.58   & -0.10$\pm$0.39   & 0.386  \\ 
  & PHQ         & -0.68$\pm$1.22   & -0.31$\pm$1.06  & 0.138  \\ 
  & SOC        & 0.81$\pm$1.10    & 0.16$\pm$0.91   & 0.060 \\                                                   
\hline
\multirow{3}{*}{Depression}                                                               
  & GAD         & 0.06$\pm$0.35   & 0.37$\pm$0.47   & \textbf{0.015}   \\ 
  & PHQ         & -0.37$\pm$1.02   & 0.30$\pm$0.73   & \textbf{0.015} \\ 
  & SOC        & 0.30$\pm$0.78   & -0.51$\pm$0.86    & \textbf{0.001} \\ 
\hline
\multirow{3}{*}{High SOC}                                                                  
  & GAD         & 0.06$\pm$0.37   & -0.27$\pm$0.47   & \textbf{0.005} \\ 
  & PHQ         & -0.31$\pm$1.00  & -0.57$\pm$1.20   & \textbf{0.037} \\ 
  & SOC        & 0.45$\pm$0.82   & 0.70$\pm$0.88   & \textbf{0.035} \\ 
\hline
\end{tabular}
\caption{Summary of intervention statistics. 
Shown are the intervention results (T1), as compared with the original results (T0), in a sample of seven \acp{llm}. 
Bold face indicates a statistically significant difference between T0 and T1, assessed by a paired t-test. }
\label{tab:intervention}
\end{table}

\section{Discussion}
\paragraph{Psychometric diagnosis:}

The evaluation of pertinent latent constructs offers a systematic method for identifying potential behavioral issues in \acp{llm}, akin to established practices in psychology. 
This study applied mental–health-related assessment tools to \acp{llm} and validated the methods and results through established techniques. 
Our findings confirm that associations known in human psychology exist in \acp{llm}.

\paragraph{Corrective interventions:}
Integrating psychological constructs into the development and testing cycle of \acp{llm} can significantly enhance our capability of understanding their behavior and improve user experience. Our results show that strengthening a positive construct, such as SoC, within \acp{llm} effectively mitigates negative psychological constructs, such as anxiety and depression.

\paragraph{\ac{nli} vs conversational prompts:}

Similar to \cite{PellertEtAl2023}, we chose \ac{nli} as an assessment method. 
Instead of using questions as premises and Likert scale options as hypotheses, the premise–hypothesis pairs should be reformulated to facilitate logical entailment with \acp{emo} inserted.

Unlike recent studies on psychometric assessment of large-scale conversational \acp{llm}, \ac{palm} is applied to base models to assess arbitrary \acp{llm}, including medium-sized and non-conversational models.
\ac{palm} mitigates some of the challenges highlighted by \cite{gupta2023investigating} and \cite{SongEtAl2023}; \ac{palm} is insensitive to questionnaire option order, unlike humans and conversational \acp{llm}.

The two-way normalization that we used to quantify biases related to the measured constructs increases the robustness of the assessment to different phrasing of prompts that convey identical concepts, as was confirmed by a high \ac{sc} and the observation that synonyms show similar trends across intensifiers.

Our framework showcases an adeptness for contextual understanding. 
On the one hand, by altering the terms related to the measured construct, we found a change in the entailment scores; on the other hand, the trends in these scores are consistent across questions that measure the same construct and are affected by contexts derived from other questions. 
The proposed method, therefore, addresses issues related to context sensitivity and reliability.

\paragraph{Fine-tuning on \ac{mnli}:} 
\label{sec:discus:freeze}
\Acp{llm} can be augmented with a new \ac{nli}, as described in \cref{sec:method:assessment}, while freezing or not freezing the weights of the base model during the fine-tuning process. 
The former option results in less accurate \ac{mnli} classifiers but leaves the base model intact, whereas the latter option results in better \ac{mnli} classifiers and reduces noise during the psychometric assessment, which, in turn, increases internal consistency (\cref{sec:method:internal-consistency}) and flexibility during prompt design (\cref{sec:methods:prompts}). 
Whereas applying the same procedure to all tested models should not affect their relative assessment, different models may react differently to fine-tuning under the same conditions, introducing unwanted biases. 
In this article, we present the results obtained without freezing the weights of the base models since we did not observe such biases during a pilot study. 
To fine-tune the models on the \ac{mnli} dataset, we used the \textit{run\_glue.py}~\footnote{\url{https://tinyurl.com/run-glue}} script provided by HuggingFace with 5e-5 learning rate and 3 epochs.

Significantly, fine-tuning the \acp{llm} to \ac{mnli} reduced both anxiety and depression scores. 
Thus, fine-tuning the models to \ac{mnli} after each domain-adaptation epoch may hinder the attribution of the changes in the measured constructs (Table \ref{tab:intervention}) to the controlled interventions. To retain validity, we fine-tuned the \ac{nli} heads once before testing the effect of the interventions.

%%%% contributions %%%%

\paragraph{Limitations and Future Work:}
Notably, \ac{palm} is unsuitable for questionnaires that measure knowledge and do not have a clear stance.
Although we paid special attention to biases introduced by fine-tuning and domain adaptation, some adverse effects may have remained unnoticed. 
Designing \ac{nli} prompts to measure latent constructs in \acp{llm} while adhering to the requirements listed in \cref{sec:methods:prompts} and avoiding caveats highlighted by related work is an arduous and time-consuming process. 
Especially challenging is the identification of \acp{emo}, intensifiers, and appropriate formulations of neutral templates while retaining the soundness of the phrases and logical entailment. 
In \cref{sec:method:challenges}, we provide examples highlighting some of the challenges. 
While automation using large-scale conversational \acp{llm} may streamline parts of the translation process, manual curation will likely remain essential, particularly for non-standardized and sensitive-topic questionnaires such as those addressing sexism.

Future research could explore \acp{llm} as proxies for the mindsets of corpus authors, building on their ability to reflect latent constructs observed in training data, akin to the virtual persona concept demonstrated by \cite{jiang2023personallm}.
Another future direction could explore how to adjust the \ac{nli} prompts and add an \ac{nli} head for conversational \ac{llm}s such as GPT and LLaMA.

\subsection{Availability}
The data and code reported in this article are publicly accessible on GitHub \url{https://github.com/cnai-lab/qlatent} under the Creative Commons license.  

\section*{Acknowledgments}
This research was partially supported by the Israeli Ministry of Science and Technology (proposal number 0005450).

\bibliographystyle{apacite} 
\bibliography{bibliography}

\appendix

\section{Background on Questionnaires}
\label{sec:questionnaires}
A questionnaire is an instrument measuring one or more constructs using aggregated item scores, called scales~\citep{oosterveld2019methods}. 
Questionnaires evolved as a research tool in the 19th century~\citep{gault1907history}, and scales are widely used to capture behavior, feelings, or actions in a range of social, psychological, and health contexts. 
These scales are based on theoretical understandings~\citep{boateng2018best} and are designed using a set of items that represent latent constructs \citep{gliem2003calculating}. 
The theoretical basis of the measured concept influences the content and structure of the questionnaire.
Therefore, the scale development process requires a thorough understanding of what we wish to measure~\citep{schrum2020four}.

\paragraph{The Likert scale}
is a widely used method in social sciences for measuring attitudes or opinions. It consists of statements that respondents rate in response to a given prompt~\citep{joshi2015likert}. 
Typically, respondents specify their level of agreement or a ranking to a particular statement; however, the use of these scales can also encompass categories, such as importance (e.g.,  from \texttt{"not important"} to \texttt{"very important"}), frequency (e.g., from \texttt{"never"} to \texttt{"always"}), and other categories~\citep{brown2010likert}. 
In this study, we created Likert scales by using existing vocabularies of intensifiers.

\paragraph{Validity} 
is a critical aspect in the development process of scales~\citep{boateng2018best}. 
An intuitive definition of validity is “…whether or not a test measures what it purports to measure”~\citep{kelley1927interpretation}. 
According to \cite{badenes2020scale}, a good validation process must address several aspects: ensuring that the scale measures the intended concept, comparing the scale with other validated measures, and ensuring that the scale does not measure unintended aspects.

\section{Main Challenges in Designing NLI Prompts}
\label{sec:method:challenges}
Below, we highlight three main challenges in transforming standard questionnaires into \ac{nli} prompts and propose a process for designing the prompts. Consider the following general structure of a question: pretext, statement, and a few responses on a Likert scale. 
We will use a question from the \ac{soc} questionnaire as a running example: \texttt{"Has it happened that people whom you counted on disappointed you?"} The answers are arranged on a 7-point Likert scale, ranging from \texttt{"never happened"} (high SoC) to \texttt{"always happened"} (low SoC). 
In all following examples, we use brackets to mark multiple options, e.g., texttt{"it [never | always] happened"} and curly braces to specify variables, e.g., \texttt{"it \{frequency\} happened"}.

Developing \ac{llm} prompts based on validated questionnaires requires careful consideration. 
The following are examples of three main challenges:

\paragraph{Congruence and linguistic acceptability:}
Consider the sentence: \texttt{"People whom I counted on encouraged disappointment."}
The phrase \texttt{"encouraged disappointment"} will receive a low probability in most \acp{llm}, regardless of any possible associations between trust and disappointment, because it is incongruent.

\paragraph{Neutrality of the template with respect to the measured construct:}
Consider the template \texttt{"Trustworthy people whom I count on [always | never] disappoint me."}
Here, the scores of \texttt{"never"} and \texttt{"always"} are extremely biased due to priming by \texttt{"trustworthy."}

\paragraph{Measuring the right thing:}
Our running example quantifies the association between trust and disappointment on a frequency scale. 
The prompt \texttt{"It happened that people whom I [never | always] counted on disappointed me"} 
is sub-optimal since the intensifiers measure the frequency of trust and not the frequency of disappointment in trusted people.

\section{List of acronyms}
\begin{acronym}[ICANN]
    %\acro  {bert}  [BERT]  {bidirectional encoder representations from transformers}
    %\acro  {gpt}   [GPT]   {generative pre-training}   
    \acro  {ai}    [AI]    {artificial intelligence}
    \acro  {xai}   [XAI]   {explainable artificial intelligence}    
    \acro  {llm}   [PLM]   {pre-trained language model}
    \acro  {nli}   [NLI]   {natural language inference}
    \acro  {mnli}  [MNLI]  {multi-genre natural language inference}
    \acro  {mlm}   [MLM]   {masked language model}
    \acro  {gad}   [GAD-7] {7-item generalized anxiety disorder}
    \acro  {phq}   [PHQ-9] {9-item patient health questionnaire}
    \acro  {soc}   [SoC-13]   {13-item Sense of Coherence}

    % \acro {palm}   [PALM]  {framework for psychometric assessment of pre-trained language models}
    \acro {palm}   [EMPALC]  {framework for evaluation of model psychometrics and assessment of latent constructs}
    \acro {emo}   [CTerm]  {term directly related to the construct being measured}
    \acrodefplural{emo}[CTerms]{terms directly related to the construct being measured}
    \acro{ss}     [SS]     {semantic similarity}
    \acro{la}     [LA]     {linguistic acceptability}
    \acro{sc}     [SC]     {silhouette coefficient}
\end{acronym}

\end{document}